\documentclass[sigconf]{acmart}

\renewcommand\footnotetextcopyrightpermission[1]{} % removes footnote with conference information in first column
\makeatletter
\renewcommand\@formatdoi[1]{\ignorespaces}
\makeatother
\AtBeginDocument{%
	\providecommand\BibTeX{{%
			\normalfont B\kern-0.5em{\scshape i\kern-0.25em b}\kern-0.8em\TeX}}}

\setcopyright{acmcopyright}
\copyrightyear{2021}
\acmYear{2021}
\acmDOI{10.1145/1122445.1122456}

\acmConference[Chengdu '2021]{Chengdu '2021: ACM International Conference on Multimedia}{October 20--24, 2021}{Chengdu, China}
\acmPrice{15.00}
\acmISBN{978-1-4503-XXXX-X/18/06}

\acmSubmissionID{1326}

\usepackage{epsfig}
\usepackage{graphicx}
\usepackage{amsmath}

\usepackage{amssymb}
\usepackage{makecell}
\usepackage{subfigure}
\usepackage{bm} % bold + italic
\usepackage{multirow} % table
\usepackage{hhline} % double lines
\usepackage{array} % equal width
\newcolumntype{M}[1]{>{\centering\arraybackslash}m{#1}} % equal width
\usepackage[utf8]{inputenc}
\usepackage[english]{babel}
\usepackage{anyfontsize} % adjust font size manually
\usepackage{mathtools} % setlength, jot

\newlength{\textfloatsepsave} 
\usepackage{algorithmicx}
\usepackage[ruled]{algorithm}
\usepackage[noend]{algpseudocode}

\usepackage{varwidth}% http://ctan.org/pkg/varwidth
\usepackage{lipsum,etoolbox}
\usepackage{graphics}

\usepackage{amssymb}% http://ctan.org/pkg/amssymb
\usepackage{pifont}% http://ctan.org/pkg/pifont
\usepackage{amsmath}
\usepackage{graphicx}
\usepackage{comment}
\usepackage{amsmath,amssymb} % define this before the line numbering.
\usepackage{color}
\usepackage{array}
\newcolumntype{C}[1]{>{\centering\arraybackslash}m{#1}}
\newcolumntype{R}[1]{>{\raggedleft\arraybackslash}m{#1}}
\newcolumntype{P}[1]{>{\raggedright\arraybackslash}p{#1}}
\newcolumntype{M}[1]{>{\centering\arraybackslash}m{#1}}
\usepackage{multirow}

\newcommand{\etal}{\textit{et al}.~}
\newcommand{\ieno}{\textit{i}.\textit{e}.}
\newcommand{\egno}{\textit{e}.\textit{g}.} %there is no space
\newcommand{\etcno}{\textit{etc}} %there is no "."

\newcommand{\tcb}{\textcolor{black}}

\newcommand{\tcbb}{\textcolor{black}}

\usepackage{enumitem}

\usepackage{xcolor}

\begin{document}
	
	%\title{Part-guided Channel-wise Feature Alignment and Completion for Occluded Person Re-Identification}
	\title{Pose-Guided Feature Learning with Knowledge Distillation \\ for Occluded Person Re-Identification}
	
    % 	\author {
    %     % Authors
    %         Kecheng Zheng\textsuperscript{\rm 1}\thanks{This work was done when Kecheng Zheng was an intern at MSRA.},
    %         Cuiling Lan\textsuperscript{\rm 2},
    %         Wenjun Zeng\textsuperscript{\rm 2},
    %         Zhizheng Zhang\textsuperscript{\rm 1},
    %         Zheng-Jun Zha\textsuperscript{\rm 1}$\thanks{Corresponding Author}$ \\
    %     }
    %     \affiliation{
    %         % Affiliations
    %         \institution{\textsuperscript{\rm 1} University of Science and Technology of China 
    %         \textsuperscript{\rm 2} Microsoft Research Asia }
    %         }
    %     \email {
    %         \{zkcys001,zhizheng\}@mail.ustc.edu.cn, \{culan, wezeng\}@microsoft.com, zhazj@ustc.edu.cn
    %         }
    
    \author{Kecheng Zheng$^{1}$, Cuiling Lan$^{2}$, Wenjun Zeng$^{2}$, Jiawei Liu$^{1}$, Zhizheng Zhang$^{2}$, Zheng-Jun Zha$^{1}$}%\footnotemark[1]
    \affiliation{%
    	\institution{$^1$University of Science and Technology of China}
    	\institution{$^2$Microsoft Research Asia} \country{}
    }\thanks{This work was done when Kecheng was an intern at MSRA. \\Corresponding authors: Cuiling Lan, Zheng-Jun Zha}
    \email{{zkcys001,jwliu6}@mail.ustc.edu.cn, {culan, wezeng,zhizzhang}@microsoft.com, zhazj@ustc.edu.cn}
    %\authornote{Corresponding authors}
%    \author{Kecheng Zheng}
%    \authornote{This work was done when Kecheng Zheng was an intern at MSRA.}
%    \email{zkcys001@mail.ustc.edu.cn}
%    \affiliation{
%        \institution{University of Science and Technology of China}
%    }
%    
%    \author{Cuiling Lan}
%    % \authornotemark[1]
%    \email{culan@microsoft.com}
%    \affiliation{
%        \institution{Microsoft Research Asia}
%    }
%    
%    \author{Wenjun Zeng}
%    \email{wezeng@microsoft.com}
%    \affiliation{
%        \institution{Microsoft Research Asia}
%    }
%    
%    \author{Jiawei Liu}
%    \email{jwliu6@ustc.edu.cn}
%    \affiliation{
%        \institution{University of Science and Technology of China}
%    }
%    
%    \author{Zhizheng Zhang}
%    \email{zhizzhang@microsoft.com}
%    \affiliation{
%        \institution{Microsoft Research Asia}
%    }
%    
%    \author{Zheng-Jun Zha}
%    % \authornote{Corresponding author.}
%    \email{zhazj@mail.ustc.edu.cn}
%    \affiliation{
%        \institution{University of Science and Technology of China}
%    }
    
	
	\begin{abstract}
	Occluded person re-identification (ReID) aims to match person images with occlusion. It is fundamentally challenging because of the serious occlusion which aggravates the misalignment problem between images. At the cost of incorporating a pose estimator, many works introduce pose information to alleviate the misalignment in both training and testing. To achieve high accuracy while preserving low inference complexity, we propose a network named Pose-Guided Feature Learning with Knowledge Distillation (PGFL-KD), where the pose information is exploited to regularize the learning of semantics aligned features but is discarded in testing. PGFL-KD consists of a main branch (MB), and two pose-guided branches, \ieno, a foreground-enhanced branch (FEB), and a body part semantics aligned branch (SAB). The FEB intends to emphasise the features of visible body parts while excluding the interference of obstructions and background (\ieno, foreground feature alignment). The SAB encourages different channel groups to focus on different body parts to have body part semantics aligned representation. To get rid of the dependency on pose information when testing, we regularize the MB to learn the merits of the FEB and SAB through knowledge distillation and interaction-based training. Extensive experiments on occluded, partial, and holistic ReID tasks show the effectiveness of our proposed network.% and validate the superiority of PGFL-KD over various state-of-the-art methods. %Specifically, our framework significantly outperforms state-of-the-art approaches by \tcr{$10.3\%$} in mAP accuracy on the challenging Occluded-Duke dataset.   
    % 	Specifically, PCAN consists of pose-enhanced module and feature alignment module. The pose-enhanced module utilizes salience pose heatmap to enhance foreground human parts towards discriminative feature learning. The feature alignment module encourages different channel groups to focus on different body parts by using interaction-based training between the global feature and body part features. This significantly improves the performance on occluded person ReID. Moreover, in order to enhance the feature of an occluded query image, we propose a graph convolutional network based on these channel-wise aligned features to borrow information from gallery images for refinement, where we learn the graph edge weights by taking into account the visibility of body joints.
    % 	Extensive experiments on occluded, partial, and holistic ReID tasks show the effectiveness of our proposed networks. Specifically, our framework significantly outperforms state-of-the-art by $7.6\%$ in mAP accuracy on the Occluded-Duke dataset.
	\end{abstract}
	
	
\begin{CCSXML}
<ccs2012>
<concept>
<concept_id>10010147.10010178.10010224.10010245.10010252</concept_id>
<concept_desc>Computing methodologies~Object identification</concept_desc>
<concept_significance>500</concept_significance>
</concept>
</ccs2012>
\end{CCSXML}

\ccsdesc[500]{Computing methodologies~Object identification}

% 	\begin{CCSXML}
% 		<ccs2012>
% 		<concept>
% 		<concept_id>10002951.10003317.10003338.10003346</concept_id>
% 		<concept_desc>Retrieval</concept_desc>
% 		<concept_significance>500</concept_significance>
% 		</concept>
% 		<concept>
% 		<concept_id>10002951.10003317.10003338.10003344</concept_id>
% 		<concept_desc>Information systems~Combination, fusion and federated search</concept_desc>
% 		<concept_significance>500</concept_significance>
% 		</concept>
% 		<concept>
% 		<concept_id>10010147.10010178.10010224.10010245.10010252</concept_id>
% 		<concept_desc>Computing methodologies~Object identification</concept_desc>
% 		<concept_significance>300</concept_significance>
% 		</concept>
% 		</ccs2012>
% 	\end{CCSXML}
	
% 	\ccsdesc[500]{Information systems~Top-k retrieval in databases}
% 	\ccsdesc[500]{Information systems~Combination, fusion and federated search}
% 	\ccsdesc[300]{Computing methodologies~Object identification}
	
    \keywords{Occluded Person Re-Identification, Human Pose, Knowledge Distillation, Feature Alignment}
	
	
	\maketitle
	
	\section{Introduction}

	\begin{figure}[t!]
		\includegraphics[width=0.95\linewidth]{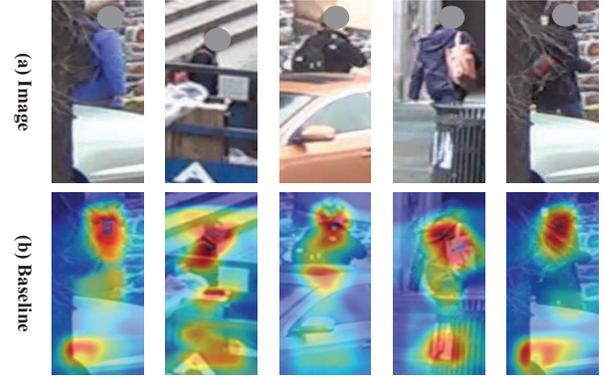}
		\caption{Examples of (a) occluded/partial person images and (b) the feature responses of Baseline. In (b), for the regions with objects occluding persons (i.e., obstructions), the networks usually mistakenly generate high responses by regarding them as discriminative person regions.}
		% and (c) our model
		%Thanks to the pose-guided feature Learning with knowledge distillation, our model using only main stream is robust to the occlusions.}% (a) Due to the occlusion/body part missing, there is feature misalignment between the holistic person and the occluded person. (b1) Decoupling of body parts features to different channel-groups (with the co-located group denoting the same body part) can alleviate this misalignment problem. (b2) The legs in the left image are occluded, but they are visible in the gallery images. We propose to borrow information from gallery images (based on the visibleness and features) for feature completion of the occluded query image.} %refinement performs semantic
		%\caption{Illustration of channel-wise aligning features and semantic feature completion. 1) Due to the occlusion, there is the feature misalignment between the holistic person and the occluded person. A channel-wise alignment feature can alleviate this effect;
		%2) The legs in the left image are occluded, but they are visible in the gallery images. Borrowing information from gallery images for refinement performs semantic feature completion of the occluded query image.}
		\label{fig:intro}
	\end{figure}
	
	Person re-identification (ReID) \cite{gong2014person,zheng2016person,zheng2020exploiting,zheng2021groupaware,Zhang_2019_CVPR,zhang2020empowering} aims to match images of a person across cameras, which has many applications such as person tracking in a retail store, finding lost child, \etcno.
	In recent years, many methods have been proposed for person ReID \cite{ma2014covariance,yang2014salient,liao2015person,zheng2013reidentification,koestinger2012large,liao2015efficient,zheng2016person,hermans2017defense,sun2018beyond,jin2020semantics,ye2021deep,zheng2020hierarchical,wang2021exploring}.
	However, most of them focus on holistic person images and only very few works investigate the more challenging occluded person ReID~\cite{zhuo2018occluded,zhuo2019novel,he2019foreground-aware, he2018deep,sun2019perceive,luo2019stnreid,miao2019PGFA,gao2020pose,wang2020high,zhai2020deep}, even though the occluded person images are very common in practical scenarios.
	As shown in Figure~\ref{fig:intro}, a person is usually occluded by some objects (\textit{e.g.} tree, car, board, bucket) or walks out of the camera field, leading to occluded or partial person images.

	Occluded/partial person ReID is challenging, where there are both occluded/partial person images and holistic person images for matching \footnote{Note that, actually, the partial person image can be also considered as the occluded person image in a broad sense where the ``occluded" region is not presented in the image.}. 
	First, the spatial misalignment between an occluded/partial person image and a holistic person image or between two occluded/partial person image is in general more severe than that between two holistic person images. 
	%As shown in Figure~\ref{fig:intro}, the legs in the left image are occluded but they are visible in the right image. Thus, the comparison w.r.t. leg regions would introduce interference/noise.
	Second, as examples shown in Figure~\ref{fig:intro}~(b), for the regions with objects occluding persons (\ieno, obstructions), the networks usually mistakenly generate high responses by regarding them as discriminative person regions, resulting in interference to the person feature representation.
	%First, the misalignment between an occluded/partial person image and a holistic person image or between two occluded/partial person image is in general more severe than that between two holistic person images. As shown in Figure~\ref{fig:intro}, the legs in the left image are occluded but they are visible in the right image. Thus, the comparison w.r.t. leg regions would introduce interference/noise. \tcp{Second, occlusion results in the missing of body part information, which is also supposed to contribute discriminative information to the feature representations.}
	
	Recently, some occluded/partial person ReID methods are proposed \cite{zhuo2018occluded,zhuo2019novel,he2019foreground-aware, he2018deep, sun2019perceive, luo2019stnreid, miao2019PGFA,gao2020pose,wang2020high}. Many works alleviate the misalignment by learning both global and local body part features \cite{miao2019PGFA,gao2020pose,wang2020high} for matching. Matching based on local body part (\egno, head, arm, leg, \etcno.) features facilitates the matching between two regions of the same semantics, alleviating the matching difficulty from spatial misalignment. Besides, such decoupling of body parts could confine the interference caused by the missing of some body parts into the local features rather than distributed to the global scope feature. On the other hand, a local body part feature may not be capable of capturing some attributes which require a more global observation (\egno, a person is wearing the clothes of the same color for upper body and lower body). \textbf{Thus, both global information and local information are vital especially for occluded person ReID. However, in general, to extract local body part features, an external pose estimator is utilized in both training and testing. This increases the complexity of the model (\ieno, model size, computational cost) in testing/inference and is not friendly in deployment.}
	
	In this work, we aim to preserve the merits of the global features and local body part features for occluded person re-identification, while eliminating the requirement of a pose estimator in testing for low complexity. To this end, we propose a Pose-Guided Feature Learning with Knowledge Distillation (PGFL-KD) network, where the pose information is exploited to regularize the learning of global features and the pose estimator is discarded in testing. PGFL-KD consists of a main branch (MB), and two pose-guided branches: a foreground-enhanced branch (FEB), and a body part semantics aligned branch (SAB). First, we explicitly alleviate the interference from the obstructions (see Figure~\ref{fig:intro}~(b)) by learning foreground-enhanced feature in the FEB, where we define the foreground as the regions around detected visible body joints based on pose. Second, based on pose, we enable different channel groups to represent features of the different body parts to have semantics aligned representation in the SAB. To get rid of the dependency on pose information when testing, the MB is ``taught" to ignore the interference from obstructions and background through knowledge distillation, and to learn semantics aligned representations through our interaction-based training, where the latter is promoted by our multi-part contrastive \tcb{loss} and interaction-based training.

	The main contributions of this paper are summarized as follows:
	
	\begin{itemize}[leftmargin=*,noitemsep,nolistsep]
		
		% \item We propose a semantic feature alignment and completion framework with the guidance of pose information. \tcb{This servers as whole system which covers feature extraction and feature retrieval enhancement.} 
		
		\item We propose a Pose-Guided Feature Learning with Knowledge Distillation (PGFL-KD) network for effective occluded person re-identification. Through pose-guided interaction learning (\ieno, knowledge distillation and interaction-based training), we enable the discarding of dependency on pose estimator while preserving high performance in testing. 
		%Part-guided Channel-wise Aligning Network (PCAN) for semantic feature alignment, which encourages different channel groups to focus on different body parts by using interaction-based training between the global feature and body part features.
		%\item We propose a Part-guided Channel-wise Aligning Network (PCAN) for semantic feature alignment, which encourages different channel groups to focus on different body parts by using interaction-based training between the global feature and body part features.
		\item We introduce two pose-guided branches in the training in order to possess two merits for teaching the MB: 1) exclusion of the interference from the obstructions and background (by the FEB); 2) semantics aligned feature representation (by the SAB). 
		
		%\item We propose a Graph Convolutional Network (GCN) based on these channel-wise aligned features to borrow information from gallery images for refinement, where we learn the graph edge weights by taking into account the visibility of body joints.

	\end{itemize}
	To the best of our knowledge, this is the first work that distills the robust feature representations based on pose in training but does not need pose in testing for occluded person ReID. Extensive experiments on occluded, partial, and holistic ReID tasks show the effectiveness of our proposed network and validate the superiority of PGFL-KD over various state-of-the-art methods. %. Specifically, our PGFL-KD significantly outperforms the state-of-the-art methods by \tcr{$\textbf{10.3\%}$} in term of mAP accuracy on the Occluded-Duke dataset. 

	\begin{figure*}[t]
		\centering
		\includegraphics[width=0.9\textwidth]{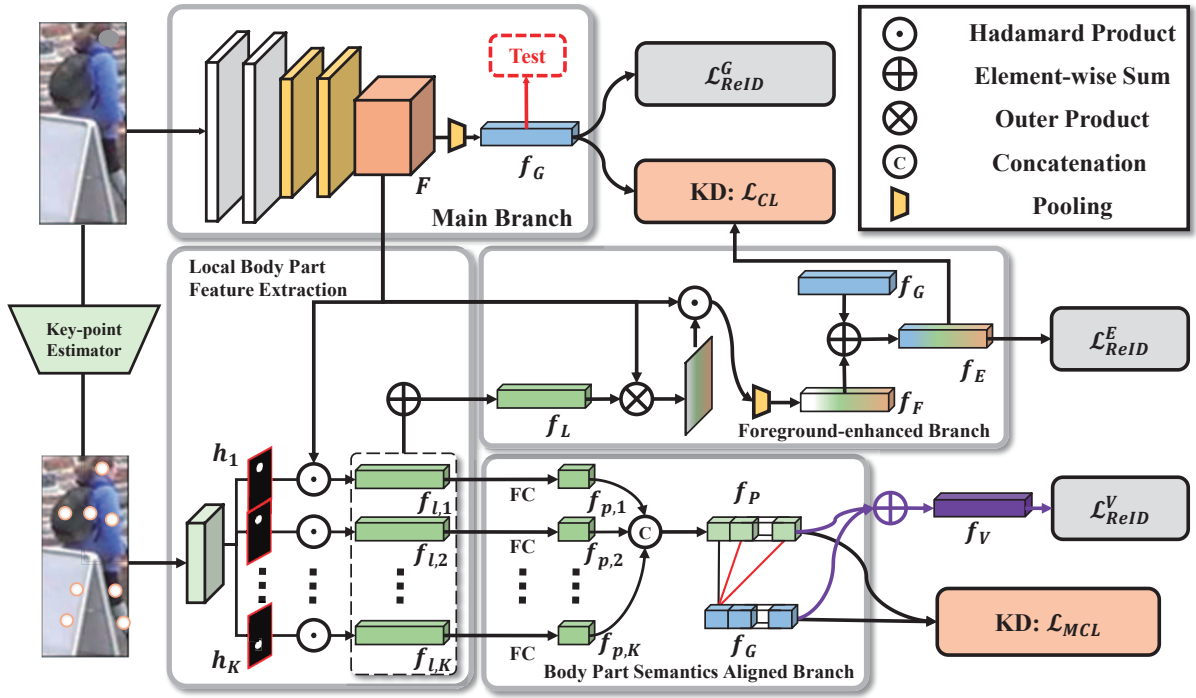}%{framework2.png}
		\caption{Illustration of our proposed network Pose-Guided Feature Learning with Knowledge Distillation (PGFL-KD), where the pose information is exploited to regularize the learning of semantics aligned features but is discarded in testing. PGFL-KD consists of a main branch (MB) and two pose-guided branches: a foreground-enhanced branch (FEB), and a body part semantics aligned branch (SAB). In testing, only the MB is needed. The FEB aims to alleviate the interference of obstructions and background by learning foreground-enhanced feature. The SAB aims to learn body part semantics aligned feature representations. We distill the knowledge from the two branches to the MB by the knowledge distillation losses (\ieno, consistent loss w.r.t. the FEB, and multi-part contrastive loss and interaction-based training (marked by purple) w.r.t. the SAB.} %channel-wise fusion optimization 
		% \caption{Illustration of our proposed semantic feature alignment and completion framework with the guidance of pose information. It consists of a basic Part-guided Channel-wise Aligning Network (PCAN), and a Graph Convolutional Network (GCN) for semantic feature completion. For PCAN, a global ReID feature vector $\boldsymbol f_{G}$ is obtained by spatially average pooling the feature map $F$. With the guidance of pose model, the body part features $\boldsymbol f_{C}$ are extracted. An interaction-based training between the global ReID feature and the local body part features encourages different channel groups of the global ReID feature to focus on different body parts \tcb{while still preserving global information}. The global ReID feature $\boldsymbol f_{G}$, body part feature $\boldsymbol f_{C}$ and the fusion feature $\boldsymbol v_{C}$ are followed by the reID losses of triplet loss and identity classification loss. Here, we enable the interaction by fusing $\boldsymbol f_{G}$ and $\boldsymbol f_{C}$, and a multi-part contrastive loss.}
		\label{fig:pipeline}
	\end{figure*}
	
	\section{Related Works}
	
	%Even though partial person images can be considered as occluded images in a broad sense, many of the previous works are specifically designed for occluded person ReID or partial person ReID and are not both suitable. Our method is suitable for both of them. In our method, we refer to ``occlusion" as the broad sense occlusion.  
	
	\noindent\textbf{Occluded/Partial Person Re-identification.}
	Occluded person ReID \cite{zhuo2018occluded} aims to match person images of both occluded/partial person and holistic person.
	Zheng \etal \cite{zheng2015partial} propose a global-to-local matching model to capture the spatial layout information. 
	He \etal \cite{he2018deep} reconstruct the feature map of a partial query from the holistic pedestrian, and further improve it with a foreground-background mask to reduce the influence of background clutter in \cite{he2019foreground-aware}.
	Iodice \etal ~\cite{iodice2018partial} align partial views by using human pose information and hallucinate the missing parts with a Cycle-Consistent Adversarial Networks \cite{zhu2017unpaired}. 
	Sun \etal propose a Visibility-aware Part Model (VPM) in \cite{sun2019perceive}, which learns to perceive the visibility of regions by self-supervised learning. 	
	Zhuo \etal \cite{zhuo2018occluded} propose occluded/non-occluded binary classification (OBC) loss \tcb{to regularize the feature learning}.
	%to distinguish the occluded images from holistic ones. %The method predicts a saliency map to highlight the discriminative parts, and utilizes a teacher-student learning scheme to further improve the learned features. 
	Luo \etal \cite{luo2019stnreid} propose a spatial transform module to transform the holistic image to align with the partial ones, and further calculate the distance of the aligned pairs.
	Fan \etal \cite{fan2018scpnet} propose a spatial-channel parallelism network (SCPNet) that encodes spatial body part features into specific channels and fuses the holistic and part features to obtain discriminative features. However, the spatial parts are obtained by dividing the feature map into several spatial horizontal stripes which cannot assure the alignment of body part semantics.  
	Recently, many works introduce a pose estimator~\cite{SunXLW19,zhang2021towards} to obtain semantics aligned local body part features, and global feature for matching \cite{miao2019PGFA,wang2020high,gao2020pose}.
	Miao \etal \cite{miao2019PGFA} propose a pose guided feature alignment method, which extracts the local body part features based on pose and alleviates the influence of occluded body regions in matching.
	%using human pose key-points to match local patches of a probe and a gallery image. They use a pre-defined threshold to determine whether a part is occluded or not based on the confidence of that key-point. 
% 	Miao \etal \cite{miao2019PGFA} propose a pose guided feature alignment method using human pose key-points to match local patches of a probe and a gallery image. They use a pre-defined threshold to determine whether a part is occluded or not based on the confidence of that key-point. 
	HOReID~\cite{wang2020high} adopts high-order relation and human-topology information for feature learning and alignment.
	PVPM~\cite{gao2020pose} utilizes the characteristic of part correspondence to estimate whether a part suffers from the occlusion or not. However, these methods extract local body part features based on pose information and a pose estimator is needed in testing, which increases the complexity of inference model.

	In this paper, we aim to exploit the local body part semantics aligned feature representations for high performance while discarding the dependency on a pose estimator in testing for low complexity.
	%Even though efforts have been made on the spatial alignment for occluded/partial Re-ID, there is still a lack of efficient optimization to enable the interaction between pose-guided features and global feature. In this paper, we propose a network named Pose-Guided Feature Learning with Knowledge Distillation (PGFL-KD), where the pose information is exploited to regularize the learning of semantics aligned features but is discarded in testing.
		
	%Even though efforts have been made on the spatial alignment for occluded/partial Re-ID, there is still a lack of efficient optimization to enable the interaction between pose-guided features and global feature. In this paper, we propose a network named Pose-Guided Feature Learning with Knowledge Distillation (PGFL-KD), where the pose information is exploited to regularize the learning of semantics aligned features but is discarded in testing.

	% but they did not consider the interaction-based training 
	
	\noindent\textbf{Knowledge Distillation.} Knowledge distillation is one of the most popular techniques in model compression and acceleration \cite{gou2021knowledge}. It in general transfers knowledge from one model (\ieno, a teacher) to another (\ieno, a student), usually from a larger model to a smaller one. In this work, we aim to transfer the knowledge from the pose-guided branches to the main branch during the training, which enables the discarding of pose-based branches while maintaining good performance.

	\section{Proposed Method}
	
    We propose a network named Pose-Guided Feature Learning with Knowledge Distillation (PGFL-KD) for occluded person ReID. Figure~\ref{fig:pipeline} shows the flowchat for training. PGFL-KD consists of three branches: a main branch (MB), a foreground-enhanced branch (FEB), and a body part semantics aligned branch (SAB). Guided by pose information, the FEB learns foreground-enhanced feature which allevates the interference from obstructions and background (see Sec. 3.2) while the SAB learns body part semantics aligned feature (see Sec. 3.3). We promote the global feature in the MB to possess the merits of the features in the other two branches by distilling knowledge from them. Particularly, we encourage the global feature $\boldsymbol f_G$ to approach the foreground-enhanced feature $\boldsymbol f_E$ by adding \tcb{consistent loss}. Moreover, we encourage the global feature $\mathbf{f}_G$ to be body part semantically aligned as $\mathbf{f}_P$ by using multi-part contrastive loss and enabling channel wise fusion. In this way, we are capable of exploiting only the MB in the testing with satisfied performance, where the pose related two branches are discarded. 
    %In Sec. 3.1, we introduce the methodology details of the Main Branch and Local Body Part Feature Extraction. 
    %In Sec. 3.2, we explicitly exclude the interference from the obstructions by learning foreground-enhanced feature in the FEB, where we define the fore-ground as the regions around detected visible body joints based on pose information. 
    %Second, in Sec. 3.3, based on pose, we enable different channel groups to represent features of the different body parts to have semantics aligned representation in the SAB.
    %Last, in Sec. 3.4, we demonstrate the formulation of the loss functions we employed in this method.
    %We will discuss the details in the following subsections.

    % 	The architecture of the proposed PGFL-KD is shown in Figure~\ref{fig:pipeline}. In our method, in order to achieve high accuracy while preserving low inference complexity, the pose information is exploited to regularize the learning of semantics aligned features but is discarded in testing. Structurally, the PGFL-KD consists of a main branch (MB), a foreground-enhanced branch (FEB), and a body part semantics aligned branch (SAB). We will discuss the details in the following subsections.

	\subsection{Feature Extraction}

	Similar to others works, we exploit both global features and local body part features in training. In contrast, we leverage the local body part features to regularize the global feature learning and only use the global feature in inference. We review how to obtain them.
	
	\noindent\textbf{Global Feature Extraction.} 
	As illustrated in Figure~\ref{fig:pipeline}, for the main branch,  we use a backbone network (\egno, ResNet-50) to extract a feature map $F \in \mathbb{R}^{h \times w \times c}$, where $h$, $w$, $c$ denote the height, width, and the number of channels, respectively. Then we adopt a global average pooling operation $g(\cdot)$ on the feature map $F$ to output a global feature $\boldsymbol f_G = g(F)\in \mathbb{R}^c$, where $c=2048$.
	%denotes the number of channels.  
	
	\noindent\textbf{Local Body Part Feature Extraction.} 
	%Previous pose-driven methods focus on tackling the large variations introduced by human pose while our method uses pose information to tackle the occlusion problems. Besides, we utilize pose key-points to encode the position information of body parts and extract local body part features.
	For local body part feature extraction, we obtain local body part features with the guidance of the estimated human pose.
	Based on the off-the-shelf human pose (key-points) estimator (HR-Net~\cite{SunXLW19}), given an input image, we obtain the heat map, with the responses identifying the estimated positions of each key point (in total 17 key points, with one channel denoting background), respectively. We merge the key-points based on semantics to have a merged heatmap $ H \in \mathbb{R}^{h\times w\times K}$ of $K=8$ channels corresponding to $K$ key-point groups: including head, left lower arm, right lower arm, left knee, right knee, left ankle, right ankle, and torso.
	%head, left and right hands, left and right feet, torso, left and right ankles. 
	We denote the $k^{th}$ channel of $H$ as $H_k = H(:,:,k)$. To suppress noise and outliers, $H_k$ is obtained by \tcb{spatially} normalizing original key-point heatmap with a softmax function. \tcb{Note that for an occluded key point, the pose estimator in general output low responses in the heatmap.}%Please refer to our supplementary for more details. 

    With the guidance of key-points regions, we get $K$ groups of semantic local features $\{\boldsymbol f_{l,k}\}|_{k=1}^{K}$ by spatially pooling the feature with each key-point heatmap as attention, respectively. We obtain the features as: 
        \begin{equation}
        \begin{aligned}
        \boldsymbol f_{l,k} = g(F \odot  (H_{k} \otimes \boldsymbol{e}_K)), \quad k = 1, \cdots, K,\\
        %\boldsymbol f_G &= g(\boldsymbol F)
        \end{aligned}
        \label{eq:local-global-features}
        \end{equation}
    where $\otimes$ denotes outer product, $\odot$ denotes Hadamard product, $g(\cdot)$ denotes global average pooling, $\boldsymbol{e}_K \in \mathbb{R}^K$ denotes a vector of all ones, $\boldsymbol f_{l,k} \in \mathbb{R}^c$.
    
    % 	With the guidance of key-points regions, we get $K$ groups of semantic local features $\{\boldsymbol f_{l,k}\}|_{k=1}^{K}$ through an outer product $\otimes$ and a global average pooling operation: 
    % 	\begin{equation}
    % 		\begin{aligned}
    % 			\boldsymbol f_L &= \{\boldsymbol f_{l,k}\}|_{k=1}^{K} = g(F \otimes \boldsymbol h_{k})|_{k=1}^{K}, \\
    % 			%\boldsymbol f_G &= g(F)
    % 		\end{aligned}
    % 		\label{eq:local-global-features}
    % 	\end{equation}
    % 	where $\boldsymbol f_{l,k} \in \mathbb{R}^c$.
    	
    Based on local body part features, we explicitly alleviate the interference from the obstructions and background by learning foreground-enhanced feature in the FEB, where we define the foreground as the regions around detected visible body joints based on pose. 
    Meanwhile, we also use the local body part features to enable different channel groups to represent features of the different body parts to have semantics aligned representation in the SAB. We introduce the FEB and SAB in details in the following subsections.

	\subsection{Foreground-enhanced Branch (FEB)}
	\label{subsec:SAB}
	
    The Foreground-enhanced Branch (FEB) intends to emphasise the features of visible body parts while excluding the interference of occluding objects and background. Specifically, we use the sum of local body part features as a query to find more salient foreground regions in the feature map to obtain enhanced foreground feature. We distill the knowledge from the FEB to the MB. 
    %and then use the enhanced feature to improve the model's perception of the foreground.
    
	\noindent\textbf{Pose-Guided Feature Enhancement.}
	With the local body part features $\boldsymbol f_{l,k}$ ($k=1,\cdots,K$) representing the informative foreground human parts, we intend to let the feature learning focus on more semantically meaningful regions. We propose a pose-guided foreground-enhanced module to improve the quality of the pooled feature by emphasising the features of visible body parts. 
    % 	With the pooled feature $\boldsymbol f_L$ being more informative about foreground human parts, we could also let the feature learning focus on more semantically meaningful representations. Inspired by recent works showing CNN’s capability of localizing salient objects, we propose a pose-guided foreground-enhanced module to improve the quality of the pooled feature by emphasising the features of visible body parts. 
	
	As illustrated in Figure~\ref{fig:pipeline}, given a feature map $F \in \mathbb{R}^{h \times w \times c}$ and its pose-based pooled feature vector $\boldsymbol f_L \in \mathbb{R}^{c}$, where $\boldsymbol f_L = (\sum_{k=1}^{K} \boldsymbol f_{l,k})/K$, we first calculate the cosine similarity between the feature $\boldsymbol f_L $ and the feature map $F$ at each pixel. Then we use softmax function to calculate the attention score map with the score value a position $(i, j)$ as 
	 \begin{equation}
        \begin{aligned}
            a_{i,j} = \frac{\exp({F_{i,j}\cdot f_L })}{\sum_{i,j} \exp({F_{i,j}\cdot f_L })}, \quad i \in [1,h], j \in [1,w],
        \end{aligned}
        \label{eq:attn}
    \end{equation}
    where $F_{i,j} \in \mathbb{R}^{c}$ denotes the feature vector at position $(i, j)$ of the feature map $F$.
    %$h$ and $w$ is the height and width of feature map $F$, and 
    After obtaining the attention score map, we use it as the weights for attentive pooling of the feature map to output the foreground feature vector as
     \begin{equation}
        \begin{aligned}
            \boldsymbol f_{F} = \sum_{i,j} a_{i,j} \times F_{i,j}.
        \end{aligned}
        \label{eq:attn}
    \end{equation}
    This attentive pooling procedure can effectively shift the focus of the pooled feature vector to the body part regions, leading to more meaningful foreground representation. In order to enable the model to preserve more complete information, we add the foreground feature and the global feature to have the foreground-enhanced feature $\boldsymbol f_E = \boldsymbol f_F + \boldsymbol f_G$.
	%This attentive pooling procedure can effectively shift the focus of the pooled feature vector to the semantically salient regions, leading to more meaningful foreground enhancement. In order to let the model preserve more compplete informative, we add the enhanced feature with the global feature $\boldsymbol f_M = \boldsymbol f_E + \boldsymbol f_G$.

	\noindent\textbf{Knowledge Distillation.} 
	In order to get rid of the dependency on pose information in testing and inherent the merit of the foreground-enhanced feature, we regularize the feature learning of the MB by distilling knowledge from the FEB using consistent loss as 
    % Specifically, we enforce the ReID feature $\boldsymbol f_G$ of ReID-Stream close to the more accurate pose enhanced feature $\boldsymbol f_M$ by minimizing the Maximum Mean Discrepancy (MMD). MMD is a distance metric to measure the domain mismatch for probability distributions. We use it to measure the discrepancy between the transformed features $\hat{r}$ and $\hat{g}$, and minimize it to drive ReID-Stream to pay more attention to human part foreground features. An empirical approximation to the MMD distance of $\boldsymbol f_G$ and $\boldsymbol f_M$ is computed as follows:
    % \begin{equation}
    %     \begin{aligned}
    %         \mathcal{L}_{CL} &= MMD({\boldsymbol f_G}, {\boldsymbol f_M}) =  \|{\mu}({\boldsymbol f_G}) - {\mu}({\boldsymbol f_M}) \|_2^2,
    %     \end{aligned}
    %     \label{eq:loss_MMD}
    % \end{equation}
    \begin{equation}
        \begin{aligned}
            \mathcal{L}_{CL} &= \|{\boldsymbol f_G} - {\boldsymbol f_E} \|_2^2,
        \end{aligned}
        \label{eq:loss_MMD}
    \end{equation} 
     where the feature $\boldsymbol f_E$ from the FEB acts as the teacher and the MB as the student (by detaching the foreground-enhanced feature $\boldsymbol f_{E}$).         % where $\mu(\cdot), \sigma(\cdot)$ denotes the calculation of mean, variance of the distributions of transformed features $\boldsymbol f_G$ and $\boldsymbol f_M$.  
	 
	\subsection{Body Part Semantics Aligned Branch (SAB)}
	\label{subsec:SAB}
	
    The SAB encourages different channel groups to focus on different body parts to have semantics aligned representation. In order to get rid of the dependency on pose information when testing, we regularize the MB to learn the merits of the SAB branches through multi-part contrastive loss for knowledge distillation and interaction-based training. 
    
    For the SAB, to generate body part semantics aligned feature representation ${\boldsymbol f}_P$, we reduce the number of dimension of each local body part feature $\boldsymbol f_{l,k}$ by $K$ to have $\boldsymbol f_{p,k}$, and concatenate them as ${\boldsymbol f}_{P} = [\boldsymbol{f}_{p,1}, \boldsymbol{f}_{p,2}, \cdots, \boldsymbol{f}_{p,K}]$.  Here $\boldsymbol f_{p,k} = ReLU(BN(W_k \boldsymbol f_{p,k}))$, where $W \in \mathbb{R}^{\frac{c}{K} \times c}$ and $BN$ denotes batch normalization operation.
    	
	\noindent\textbf{Multi-part Contrastive Loss.}
	We use the multi-part contrastive loss to explicitly align the global feature $\boldsymbol{f}_{G}$ with the local body part feature $\boldsymbol{f}_{P}$ to have the semantics aligned feature representation. 
	
	To align with the local part feature $\boldsymbol{f}_{P}$, we split the global feature $\boldsymbol{f}_{G}$ of the MB into $K$ groups, \ieno, $\boldsymbol{f}_{G}=[\boldsymbol{f}_{g,1}, \boldsymbol{f}_{g,2}, \cdots, \boldsymbol{f}_{g,K}]$, where $\boldsymbol{f}_{g,k} \in \mathbb{R}^{c/K}$. Particularly, for an input image, we encourage the consistency of the features between a local part feature $\boldsymbol{f}_{p,k}$ and its corresponding channel-group $\boldsymbol{f}_{g,k}$ of the global feature, and encourage the dissimilarity of the features between a local part $\boldsymbol{f}_{p,k}$ and a channel-group $\boldsymbol{f}_{g,i}$ of a different body part of the global feature, where $i \neq j$. To exploit the symmetry of a human body, we consider that feature of the left part (\egno, left shoulder) of the body should also be close to the feature of the corresponding right part (\egno, right shoulder). By following the design of the multi-positive contrastive loss~\cite{han2020self}, for an image, we have the multi-part contrastive loss as
	\begin{equation}
	\small
	\mathcal{L}_{MCL}=-\sum_{i=1}^{K} \log \frac{\sum_{j \in \mathcal{P}(i)} \exp{( \boldsymbol f_{g,j} \cdot \boldsymbol f_{p,i} )}}{\sum_{j \in \mathcal{P}(i)} \exp{(\boldsymbol f_{g,j}  \cdot \boldsymbol f_{p,i} )}+\sum_{j \in \mathcal{N}(i)} \exp{(\boldsymbol f_{g,j} \cdot \boldsymbol f_{p,i})}},
	\end{equation}
	where $\mathcal{N}(i)$ denotes the negative set within the global feature $\boldsymbol f_{G}$ \textit{w.r.t.} the $i^{th}$ part feature $\boldsymbol f_{p,i}$. 
	For example, when the $i^{th}$ part feature $\boldsymbol f_{p,i}$ denotes the feature of left foot, the left foot and right foot features in the global feature belongs to positive set while other part features belongs to negative set. 
	%the positive sample set of global feature $\boldsymbol f_{p,2}$ is the aligned feature of left hand ($\boldsymbol f_{g,2}$) and aligned feature of right hand ($\boldsymbol f_{g,3}$). The positive set $\mathcal{P}(i)$ is the complement of the negative set. Our designed multi-part contrastive loss considers the symmetry of a human body. 
	Note that in such distillation, the local body part features act as teacher and the MB acts as student (by detaching the local body part feature $\boldsymbol f_{P}$).    %of corresponding positions of the global feature    

\noindent\textbf{Interaction-based Optimization.} Besides the above multi-part contrastive loss for distilling semantics aligned representation, we enable the interaction between local part features and global feature to promote the channel-wise semantic alignment through joint learning by fusing.
	
%To better explore the channel-wise feature alignment, we explicitly align the split channel groups of global feature and body part features together. We reduce the dimension of each semantic local feature by a mapping function \ieno, $\boldsymbol{f}_{a,k} = {\rm{ReLU}}(W \boldsymbol{f}_{l,k}) \in \mathbb{R}^{d}$, where $W \in \mathbb{R}^{d\times c}$, $d=c/K$, $k=1,\cdots,K$.
%We concatenate the $K$ local features to have a part-aligned local feature vector $\boldsymbol{f}_P = [\boldsymbol{f}_{p,1}, \boldsymbol{f}_{p,2}, \cdots, \boldsymbol{f}_{p,K}] \in \mathbb{R}^c$.
	    
% 	\noindent\textbf{Joint Optimization with Feature Interaction.} We enable the interaction between local features and global feature and promote the channel-wise semantic alignment through joint learning by fusing.
	
% 	To better explore the channel-wise feature alignment, we explicitly align the split channel groups of global feature and body part features together. We reduce the dimension of each semantic local feature by a mapping function \ieno, $\boldsymbol{f}_{a,k} = {\rm{ReLU}}(W \boldsymbol{f}_{l,k}) \in \mathbb{R}^{d}$, where $W \in \mathbb{R}^{d\times c}$, $d=c/K$, $k=1,\cdots,K$.
% 	We concatenate the $K$ local features to have a part-aligned local feature vector $\boldsymbol{f}_P = [\boldsymbol{f}_{p,1}, \boldsymbol{f}_{p,2}, \cdots, \boldsymbol{f}_{p,K}] \in \mathbb{R}^c$.

% 	\emph{Two-Branch Fusion:}
	%We enable the joint optimization of the two branches by fusing the global features with the local part features by element-wise addition, {\it{i.e.}}, 
	Particularly, we enable the joint optimization of the two branches by fusing the global features with the local part features by element-wise addition, {\it{i.e.}}, 	
	\begin{equation}
		\begin{aligned}
			\boldsymbol{f}_V = \boldsymbol{f}_{G} + \boldsymbol{f}_{P}.
		\end{aligned}
		\label{eq:local-global-features}
	\end{equation}	
	The widely-used ReID loss $\mathcal{L}_{ReID}$ ({\it{i.e.}}, the cross-entropy loss for identity classification (ID Loss), and the ranking loss of triplet loss with batch hard mining \cite{hermans2017defense} (Triplet Loss)), is added on the fused feature $\boldsymbol{f}_{V}$, which we refer to $\mathcal{L}_{ReID}^{V}$. 
	
	The fusion followed by supervision plays the role of assisting the feature alignment, which drives different channel groups of the global feature to focus on different human body parts. We will give the analysis below.

	\emph{Analysis from Perspective of Gradients:}
	For the body part feature $\boldsymbol{f}_P$, features of different local body parts are allocated into different channel groups of $\boldsymbol{f}_P$ and are thus semantically aligned across two images. Global feature $\boldsymbol{f}_G$ is not naturally semantically aligned but it contains more comprehensive information. We promote their interaction by element-wisely fusing them. We analyze the optimization gradients for the two branches below. 
	
	We take the triplet-loss as an example to analyze the gradients for the two branches, where the analysis w.r.t the classification loss is similar. We denote  the two branch fused features of an anchor sample, a positive sample, and a negative sample as $\boldsymbol{v}_{a} = \boldsymbol{f}_{G}^a + \boldsymbol{f}_{P}^a$, $\boldsymbol{v}_{p} = \boldsymbol{f}_{G}^p + \boldsymbol{f}_{P}^p$, $\boldsymbol{v}_{n} = \boldsymbol{f}_{G}^n + \boldsymbol{f}_{P}^n$ respectively (that could be sampled from a mini-batch). 
	
	We define the triplet loss on the positive-pair and the negative-pair as
	\begin{equation}
		\begin{aligned}
			\mathcal{L}_{tri}=-\log \frac{e^{\boldsymbol v_{a}^{T} \cdot \boldsymbol v_{p}}}{e^{\boldsymbol v_{a}^{T} \cdot \boldsymbol v_{p}}+e^{\boldsymbol v_{a}^{T} \cdot \boldsymbol v_{n}}}=-\log \left(1+e ^{\boldsymbol v_{a}^{T} \cdot \boldsymbol v_{n} - \boldsymbol v_{a}^{T} \cdot \boldsymbol v_{p}}\right).
		\end{aligned}
	\end{equation}
	For the triplet loss, the gradients \textit{w.r.t.} the two features are as
	\begin{equation}
		\begin{aligned}
			\frac{\partial \mathcal L_{tri}}{\partial \boldsymbol f_{G}^a}&=\frac{\partial \mathcal L_{tri}}{\partial \boldsymbol v_{a}} \cdot \frac{\partial \boldsymbol v_{a}}{\partial \boldsymbol f_{G}^a}
			=\frac{\boldsymbol v_p - \boldsymbol v_n}{1+e^{(\boldsymbol f_{G}^a + \boldsymbol f_{P}^a)^{T} \cdot \boldsymbol v_p - (\boldsymbol f_{G}^a + \boldsymbol f_{P}^a)^{T} \cdot \boldsymbol v_n}},
		\end{aligned}
	\end{equation}
	\begin{equation}
		\begin{aligned}
			\frac{\partial \mathcal L_{tri}}{\partial \boldsymbol f_{P}^a}&=\frac{\partial \mathcal L_{tri}}{\partial \boldsymbol v_{a}} \cdot \frac{\partial \boldsymbol v_{a}}{\partial \boldsymbol f_{P}^a} %\\
			=\frac{\boldsymbol v_p - \boldsymbol v_n}{1+e^{(\boldsymbol f_{G}^a + \boldsymbol f_{P}^a)^{T} \cdot \boldsymbol v_p - (\boldsymbol f_{G}^a + \boldsymbol f_{P}^a)^{T} \cdot \boldsymbol v_n}}.
		\end{aligned}
	\end{equation}
	We can see that the gradient for each branch/feature is related with/influenced by the feature of the other branch, which denotes they are not independent but interacted. 
	%The global feature $\boldsymbol{f}_G$ is guided to be semantically aligned.
	%For the body part feature $\boldsymbol{f}_P$, the gradients of the aligned features $\boldsymbol{f}_P$ reinforces $\boldsymbol{f}_G$ for semantic feature alignment. 
	%Moreover, global feature $\boldsymbol{f}_G$ contains more comprehensive information which can bring additional information to the body part feature $\boldsymbol{f}_P$.
	Moreover, the optimization direction (gradient) \textit{w.r.t.} the global feature and that \textit{w.r.t.} the part-aligned local feature are the same. When their optimization directions are the same, the two features share similar behaviors and are prone to have consistent characteristics/semantics. Specifically, the local feature is semantically aligned and thus encourages the global feature to be similarly semantically aligned.	
    % 	The channel-wise addition between the global feature $\boldsymbol{f}_G$ and the part-aligned local feature $\boldsymbol{f}_P$, and the optimization losses on the sum make the optimization direction (gradient) \textit{w.r.t.} the global feature and that \textit{w.r.t.} the part-aligned local feature the same. When their optimization directions are the same, the two features share similar behaviors and are prone to have consistent characteristics/semantics. Specifically, the local feature is semantically aligned and thus encourages the global feature to be similarly semantically aligned.    

	\subsection{Overall Loss Function}
	To drive both global branch and local branch to learn discriminative feature representations, we add the ReID loss on the global feature $\boldsymbol{f}_{G}$ (denoted as $\mathcal{L}_{ReID}^G$), the foreground-enhanced feature $\boldsymbol{f}_{E}$ (denoted as $\mathcal{L}_{ReID}^E$), and the body part semantics aligned feature (after fusion) $\boldsymbol{f}_{V}$ (denoted as $\mathcal{L}_{ReID}^V$). Together with the distillation losses, the overall loss is as
	\begin{equation}
		\mathcal L = \mathcal{L}_{ReID}^G + \mathcal{L}_{ReID}^E + \mathcal{L}_{ReID}^V  + \lambda_{cl}\mathcal{L}_{CL} + \lambda_{mcl}\mathcal L_{MCL},
		%\mathcal L_{S} + \lambda_{mc} \
	\end{equation}
	where $\lambda_{cl}$ and $\lambda_{mcl}$ denote hyper-parameters for balancing the losses. 

	\subsection{Inference/Testing}	
	In the testing phase, we discard the pose estimator and only use the main branch (MB), where the feature $\boldsymbol f_G$ is used for matching. This removes the dependency on a pose estimator and makes the model simple with low computational complexity in testing.

	\section{Experiments}

	\subsection{Datasets and Evaluation Metrics}
	
	We evaluate our methods using four person ReID datasets, including two occluded datasets (Occluded-Duke~\cite{miao2019PGFA}, and Occluded-ReID~\cite{zhuo2018occluded}), three partial datasets (Partial-REID~\cite{he2018deep}, Partial-iLIDS~\cite{he2018deep}, and our generated Partial-Duke), and two holistic datasets (DukeMT
	MC-reID~\cite{ristani2016performance} and Market-1501~\cite{zheng2015scalable}), with details shown in Table \ref{tab:datasets}. 
	
	\noindent\textbf{Occluded Person ReID Datasets.}
	These datasets focus more on occluded person images, where in a cropped person image, a person is usually occluded by some other objects/obstructions.
	Occluded-Duke ~\cite{miao2019PGFA} is generated from DukeMTMC-reID by leaving occluded images and filtering out some noisy images.
	It contains 15,618 training images, 17,661 gallery images, and 2,210 occluded query images.
	Occluded-ReID ~\cite{zhuo2018occluded} is captured by the mobile camera, consisting of 2000 images of 200 occluded persons. Each identity has five full-body person images and five occluded person images with different types of severe occlusions.
	
	\noindent\textbf{Partial Person ReID Datasets.} These datasets focus more on partial person images, where only a partial of a person is presented in the image due to imperfect detection or out of camera field. In a broad sense, these are also occluded person images.
	Partial-REID ~\cite{he2018deep} includes 600 images from 60 people, with five full-body images and five partial images per person, which is only used for the test.
	Partial-iLIDS ~\cite{he2018deep} is based on the iLIDS~\cite{he2018deep} dataset and contains a total of 238 images from 119 people captured by multiple non-overlapping cameras in the airport, and their occluded regions are manually cropped.
	
	The existing partial person ReID datasets are too small for reliable training and testing. For example, Partial-REID \cite{he2018deep} includes only 600 images from 60 people and Partial-iLIDS \cite{he2018deep} includes only 238 images from 119 people. There is a lack of large size partial person ReID dataset. To facilitate the investigation and evaluation, we generate a large partial person ReID dataset based on DukeMTMC-reID. We refer to it as Partial-Duke.
    The original DukeMTMC-reID dataset is not designed for the investigation/evaluation of partial person ReID due to its small number of partial person images. \emph{We manually generate the Partial-Duke dataset}. 
    Partial-Duke contains 50\% partial images and 50\% holistic images for the training/query/gallery sets. For these partial images, a half of them are the cropped upper half (prone to be the upper body) of the original images, and another half of images are the cropped upper one third of the original images (prone to be the upper body). In total, it contains 702 identities of 16,522 training images, 702 identities of 2,228 queries, and 1110 identities of 17,661 gallery images. 
    %We will release this dataset.  
	
	\begin{table}[t]
		\begin{center}
			\small
			\label{tab:datasets}
			\scalebox{0.97}{
				\begin{tabular}{c|c|c|c}
					\hline
					\hline
					\multicolumn{1}{c|}{\multirow{2}{*}{Dataset}} & \multicolumn{1}{c|}{\multirow{2}{*}{\begin{tabular}[c]{@{}c@{}}Train Nums\\ (ID/Image)\end{tabular}}} & \multicolumn{2}{c}{Testing Nums (ID/Image)}                \\ \cline{3-4} 
					\multicolumn{1}{c|}{}                         & \multicolumn{1}{c|}{}                                                                                   & \multicolumn{1}{c|}{Gallery} & \multicolumn{1}{c}{Query} \\ \hline
					
					Market-1501~\cite{zheng2015scalable}       & 751/12,936 & 750/19,732 & 750/3,368 \\
					DukeMTMC-reID~\cite{ristani2016performance}         & 702/16,522 & 1,110/17,661 & 702/2,228 \\ \hline
					Occluded-Duke~\cite{miao2019PGFA}   & 702/15,618 & 1,110/17,661 & 519/2,210 \\ 
					Occluded-ReID~\cite{zhuo2018occluded}   & -      & 200/1,000  & 200/1,000 \\
					\hline
					Partial-REID~\cite{he2018deep}    & -      & 60/300    & 60/300 \\
					Partial-iLIDS~\cite{he2018deep}   & -      & 119/119   & 119/119 \\ 
					Partial-Duke (Ours)   &  702/16,522 & 1,110/17,661 & 702/2,228 \\
					\hline
					\hline
				\end{tabular}
			}
		\end{center}
		\caption{Dataset details. We evaluate our proposed method on seven public datasets, including two occluded datasets, three partial datasets and two holistic ones.}
		\label{tab:datasets}
		\vspace{-10pt}
	\end{table}

	\noindent\textbf{Holistic Person ReID Datasets.}
	Market-1501 ~\cite{zheng2015scalable} and DukeMT
	MC-reID ~\cite{ristani2016performance} sare two widely-used large-scale holistic ReID datasets.
	Market-1501 contains 1,501 identities captured from 6 cameras. It has 19,732 gallery images, and 12,936 training images. This dataset contains very few occluded or partial person images.
	DukeMTMC-reID dataset contains 1,404 identities, 16,522 training images, 2,228 queries, and 17,661 gallery images.
	
	\noindent\textbf{Evaluation Metrics}. We use standard metrics in most person ReID literature, namely Cumulative Matching Characteristic Rank-1/5/10 (\ieno, R1/R5/R10) and mean average precision (mAP).%, to evaluate the quality of different person ReID models. %All the experiments are performed in a single query setting.

    \subsection{Implementation Details}
	
	For our PGFL-KD, We use ResNet50 pre-trained on ImageNet~\cite{russakovsky2015imagenet} as our backbone network. Similarly, we build our baseline scheme Baseline using ResNet50. As ~\cite{luo2019bag}, we perform data augmentation of randomly erasing~\cite{zhong2017random}, cropping, and flipping. The images are resized to $384 \times 128$. Each mini-batch contains 64 images of 4 identities, where there are 16 images for each identity. Adam~\cite{kingma2014adam} optimizer is adopted to optimize the networks. The initial learning rate is set to 0.00035. For the identity classifiers, a BNNeck is adopted, which contains a batch normalization layer~\cite{ioffe2015batch}, and a fully connected layer followed by a softmax function. The network is jointly trained end-to-end for 120 epochs with an initialized learning rate of 3.5e-4. The learning rate is decayed by 0.1 at 30 and 70 epochs. We implement our framework with Pytorch. 
	
	The HR-Net~\cite{SunXLW19} trained on the COCO dataset~\cite{lin2014microsoft} is used to extract the human key-points. The keypoint extractor predicts 17 key-points, and we merge these key-points according to the body semantics to obtain $K=8$ key-points. Specifically, torso consists of left/right shoulders and hips. We merge the left (or right) elbow and wrist as the left (or right) lower arm. After merging, the $K=8$ key-points consist of head, left lower arm, right lower arm, left knee, right knee, left ankle, right ankle, and torso.
	%Other margin manners can also be design choices. 
	% while assigning the shoulder to be part of the torso
	%head, left, and right hands, left and right feet, torso, shoulders, left and right ankles.

% 	We evaluate our methods using four person ReID datasets, including two occluded datasets (Occluded-Duke~\cite{miao2019PGFA}, and Occluded-ReID~\cite{zhuo2018occluded}), three partial datasets (Partial-REID~\cite{he2018deep}, Partial-iLIDS~\cite{he2018deep}, and \tcr{our generated} Partial-Duke), and two holistic datasets (DukeMTMC-reID~\cite{ristani2016performance} and Market-1501~\cite{zheng2015scalable}), with details shown in Table \ref{tab:datasets}. 

	\begin{table}[t]
		\centering
		\small
		\begin{tabular}{l|cc|cc}
			\hline 
			\hline 
			\multicolumn{1}{l|}{\multirow{2}{*}{Methods}} & \multicolumn{2}{c|}{Occluded-Duke}            & \multicolumn{2}{c}{Occluded-REID}       \\ 
			\multicolumn{1}{c|}{}                                 & \multicolumn{1}{c}{Rank-1} & \multicolumn{1}{c|}{mAP} & \multicolumn{1}{c}{Rank-1} & \multicolumn{1}{c}{mAP} 
			\\ \hline 
			LOMO+XQDA~\cite{liao2015person}  & 8.1 & 5.0 & - & - \\
			DIM~\cite{yu2017devil} & 21.5 & 14.4 & - & -\\
			Part-Aligned~\cite{zhao2017deeply} & 28.8 & 20.2 & - & -\\
			HACNN~\cite{li2018harmonious}  & 34.4 & 26.0 & - & -\\
			Random Erasing \cite{zhong2017random}  & 40.0 & 30.0 & - & - \\
			PCB~\cite{sun2018beyond}  & 42.6 & 33.7 & 41.3 & 38.9 \\ 
			AFPB\cite{zhuo2018occluded}  & - & - & 68.2 & - \\ 
			\hline
			Part Bilinear~\cite{suh2018part}  & 36.9 & - & - & - \\ 
			FD-GAN~\cite{ge2018fd}  & 40.8 & - & - & - \\ 
			\hline
			AMC+SWM~\cite{zheng2015partial}  & - & - & 31.2 & 27.3 \\
			DSR~\cite{he2018deep} & 40.8 & 30.4 & 72.8 & 62.8 \\
			SFR~\cite{he2018recognizing}  & 42.3 & 32 & - & - \\
			\hline
			Ad-Occluded~\cite{huang2018adversarially} & 44.5 & 32.2 & - & - \\
			TCSDO~\cite{zhuo2019novel} & - & - & 73.7 & 77.9 \\ 
			FPR~\cite{he2019foreground-aware}  & - & - & 78.3 & 68.0 \\
			\tcbb{PGFA w/o pose~\cite{miao2019PGFA}}  & 46.0 & 34.4 & - & - \\ 
			PGFA~\cite{miao2019PGFA}  & 51.4 & 37.3 & - & - \\ 
			PVPM~\cite{gao2020pose}  & - & - & 66.8 & 59.5 \\ 
			PVPM+Aug~\cite{gao2020pose}  & - & - & 70.4 & 61.2 \\ 
			HOReID~\cite{wang2020high} &  55.1& 43.8& 80.3& 70.2 \\
            ISP*~\cite{zhu2020identity} &62.8& 52.3& - & -   \\
			\hline
			{Baseline} &52.7&45.9&   73.6          &61.5  \\
			\textbf{PGFL-KD (Ours)}              &  \textbf{63.0} & \textbf{54.1} & \textbf{80.7} & \textbf{70.3}\\
			%  \textbf{} (\textit{Ours})              &  \textbf{75.7} & \textbf{74.3} & \textbf{82.3} & \textbf{72.4} \\
			\hline     
			\hline   
		\end{tabular}
		\caption{Comparison with state-of-the-arts on two occluded datasets, \textit{i.e.} Occluded-Duke~\cite{miao2019PGFA} and Occluded-REID~\cite{zhuo2018occluded}. * denotes that ISP~\cite{zhu2020identity} uses HRNet-W32 as the backbone \tcbb{and all other methods use ResNet50 backbone}.}%
		\vspace{-8mm}
		\label{tab:occluded_result}
	\end{table}

	\subsection{Comparison with the State-of-the-Arts}
	
	\noindent\textbf{Results on Occluded Person ReID Datasets.} 
	As are shown in Table~\ref{tab:occluded_result}, we mainly compare with methods of four categories: vanilla holistic ReID methods \cite{zhao2017deeply,sun2018beyond}, holistic ReID methods with key-point information \cite{suh2018part,ge2018fd}, partial ReID methods \cite{zheng2015partial,he2018deep,he2018recognizing}, and occluded ReID methods \cite{huang2018adversarially,zhuo2019novel,he2019foreground-aware,miao2019PGFA}. 
	
	The first two category approaches achieve less satisfactory results, because they do not design the networks specific to the occluded ReID. 
	For partial/occluded ReID methods, an obvious improvement is achieved on the two datasets.
	Our proposed PGFL-KD achieves the best performance when compared with these state-of-the-art methods, which outperforms the second best method HOReID \cite{wang2020high} by \textbf{10.3\%} in mAP accuracy on the large dataset Occluded-Duke. Note that HOReID needs a pose estimator in testing but we do not. 
	At the same inference complexity, our PGFL-KD outperforms the baseline scheme Baseline significantly by \textbf{8.2\%} and \textbf{8.8\%} in mAP on Occluded-Duke and Occluded-REID, respectively.
	%Note that for Occluded-REID, there is less improvement space on it since it is very small and less challenging, where its performance is already high accuracy. It demonstrates that semantic feature completion can complete occluded features to help find more good matching results.

	\begin{table}[t]
	\begin{center}
	\scalebox{1}{
	\begin{tabular}{l|cc|cc}
	\hline
	\hline 
	\multicolumn{1}{l|}{\multirow{2}{*}{Methods}} & \multicolumn{2}{c|}{Partial-REID}            & \multicolumn{2}{c}{Partial-iLIDS}       \\ 
	\multicolumn{1}{c|}{}                                 & \multicolumn{1}{c}{Rank-1} & \multicolumn{1}{c|}{Rank-3} & \multicolumn{1}{c}{Rank-1} & \multicolumn{1}{c}{Rank-3} \\ \hline
	 % AMC+SWM \cite{zheng2015partial} & 37.3 & 46.0 & 21.0 & 32.8 \\
	 DSR \cite{he2018deep}& 50.7 & 70.0 & 58.8 & 67.2 \\
	 SFR \cite{he2018recognizing} & 56.9 & 78.5 & 63.9 & 74.8 \\
	 VPM \cite{sun2019perceive} & 67.7 & 81.9 & 65.5 & 74.8 \\ 
	 PGFA \cite{miao2019PGFA} & 68.0 & 80.0 & 69.1 & 80.9 \\
	 AFPB \cite{zhuo2018occluded} & 78.5 & - & - & - \\ 
	 FPR \cite{he2019foreground-aware} & 81.0 & - & 68.1 & - \\ 
	 TCSDO \cite{zhuo2019novel}& 82.7 & - & - & - \\ 
	 HOReID~\cite{wang2020high}  & \textbf{85.3} & \textbf{91.0} & {72.6} &{86.4}\\

	 \hline
	 \textbf{PGFL-KD (Ours)} & {85.1} & {90.8} & \textbf{74.0} &\textbf{86.7}\\
	\hline
	\hline
	\end{tabular}
	}
	\end{center}
	\caption{Comparison with state-of-the-art approaches on two partial datasets, \textit{i.e.} Partial-REID~\cite{zheng2015partial} and Partial-iLIDS~\cite{he2018deep} datasets. Our method achieves competitive performance on the two partial datasets.}
	\label{tab:partial_result}
	\vspace{-6mm}
	\end{table}

	\begin{table}[t]%[h]
		\centering
		\small
		\begin{tabular}{l|cccc}
			\hline 
			\hline 
			\multicolumn{1}{l|}{\multirow{2}{*}{Methods}} & \multicolumn{4}{c}{Partial-Duke}       \\ 
			\multicolumn{1}{c|}{}                                 & \multicolumn{1}{c}{Rank-1} & \multicolumn{1}{c}{Rank-5} & \multicolumn{1}{c}{Rank-10} & \multicolumn{1}{c}{mAP} 
			\\ \hline 
			
			FPR~\cite{he2019foreground-aware}  & 69.2  &83. 4& 87.6   &  50.5      \\
			PGFA~\cite{miao2019PGFA}  & 66.2  &81.5 & 85.4   &  42.5           \\ 
			PVPM~\cite{gao2020pose}  & 74.6  &  83.7&   88.9&  57.3 \\ 
			HOReID~\cite{wang2020high} &  77.6  &  86.3&   90.9&  59.0   \\
			\hline
			Baseline              &  {70.1} & {82.2} & {87.7} & {51.2}\\
			\textbf{PGFL-KD (Ours)}              &  \textbf{81.1} & \textbf{89.5} & \textbf{92.7} & \textbf{64.2}\\
			\hline   
		\end{tabular}
		\caption{Performance comparison (\%) with the state-of-the-arts on our created large partial dataset, \textit{i.e.} Partial-Duke. }
		%PGFL-KD (Ours) achieves best performance.
		\label{tab:partialreid_result}
		\vspace{-6mm}
	\end{table}
	\noindent \textbf{Results on Partial Person ReID Datasets.}
	%
	%Accompanied by occluded images, partial ones often occur due to imperfect detection, outliers of camera views, and so on.
	To further evaluate our proposed scheme, in Table \ref{tab:partial_result} we report the results on two partial person ReID datasets, Partial-REID~\cite{zheng2015partial} and Partial-iLIDS~\cite{he2018deep}.
	As we can see, our proposed PGFL-KD outperforms the other methods by at least $1.4\%$ in terms of Rank-1 accuracy on Partial-iLIDS and achieves the competitive results to HOReID\cite{hao2019hsme} on Partial-REID. Our inference model is simple and does not need pose estimator but HOReID requires.
	
	The existing partial person ReID datasets are too small for reliable training and testing. Thus we manually generate the Partial-Duke dataset, which is much larger than Partial-REID and Partial-iLIDS (see Table \ref{tab:datasets}. Table \ref{tab:partialreid_result} shows the comparison with the state-of-the-art approaches on this large Partial-Duke dataset, where the results are obtained by running their source codes. We can see that our proposed PGFL-KD achieves the best performance, which outperforms the second best method by \textbf{5.2\%} in mAP accuracy.

	\begin{table}[t]
		\begin{center}
			\scalebox{1}{
				\begin{tabular}{l|cc|cc}
					\hline
					\hline
					\multicolumn{1}{l|}{\multirow{2}{*}{Methods}} & \multicolumn{2}{c|}{Market-1501} & \multicolumn{2}{c}{DukeMTMC} \\ 
					\multicolumn{1}{c|}{} & \multicolumn{1}{c}{Rank-1} & \multicolumn{1}{c|}{mAP} & \multicolumn{1}{c}{Rank-1} & \multicolumn{1}{c}{mAP} \\ 
					\hline 
					PCB \cite{sun2018beyond} & 92.3 & 77.4 & 81.8 & 66.1 \\
					VPM \cite{sun2019perceive} & 93.0 & 80.8 & 83.6 & 72.6 \\  
					BOT \cite{luo2019bag} & 94.1 & 85.7 & 86.4 & 76.4 \\
					GCP~\cite{park2020relation}&95.2 & \textbf{88.9}&87.9 & 78.6  \\
				    %RGA~\cite{zhang2020relation} & \textbf{96.1} &88.4& - & - \\
					\hline
					SPReID \cite{kalayeh2018human}& 92.5 & 81.3 & - & - \\
					MGCAM \cite{song2018mask} & 83.8 & 74.3 & 46.7 & 46.0 \\
					MaskReID \cite{qi2018maskreid} & 90.0 & 75.3 & - & -\\
                    ISP~\cite{zhu2020identity} &-&-&{88.7} &{78.9}  \\
                    % ISP(HRNet)~\cite{zhu2020identity} &95.3& 88.6&\textbf{89.6} &\textbf{80.0}  \\
				    % 	FPR \cite{he2019foreground-aware} & 95.4 & 86.6 & 88.6 & 78.4 \\ 
					\hline
					PDC \cite{su2017pose} &  84.2 & 63.4 & - & - \\
					Pose-transfer \cite{liu2018pose} & 87.7 & 68.9 & 30.1 & 28.2 \\
					PSE \cite{saquib2018pose} & 87.7 & 69.0 & 27.3 & 30.2 \\
					PGFA \cite{miao2019PGFA} & 91.2 & 76.8 & 82.6 & 65.5 \\ 
					HOReID~\cite{hao2019hsme} & 94.2 & 84.9 & 86.9 & 75.6 \\
                    GASM~\cite{he2020guided} &\textbf{95.3}& 84.7&88.3 & 74.4  \\
					\hline
					Baseline & 94.0 & 85.2 & 86.3 & 76.1 \\
					\textbf{PGFL-KD (Ours)}&\textbf{95.3}&{87.2}&\textbf{89.6}&\textbf{79.5}\\
					
					\hline
					\hline
				\end{tabular}
			}
		\end{center}
		\caption{Comparison with state-of-the-arts on two holistic datasets, Market-1501 and DukeMTMTc-reID.}
		\vspace{-6mm}
		\label{tab:holistic_results}
		% \cite{ristani2016performance,zheng2017unlabeled}
		% Our method achieves competitive performance.
	\end{table}

	\noindent\textbf{Results on Holistic Person ReID Datasets.}
	In considering the practical applications where both occluded and holistic person matching is needed, it is expected that a method designed for occluded person ReID should work for holistic person ReID. We compare with the state-of-the-art approaches on holistic person ReID in Table \ref{tab:holistic_results}. We also compare with the vanilla ReID methods \cite{sun2018beyond,sun2019perceive,luo2019bag}, the ReID methods with human-parsing information \cite{kalayeh2018human,song2018mask,qi2018maskreid,zhu2020identity}, and the holistic ReID methods with key-points information \cite{su2017pose,liu2018pose,saquib2018pose,miao2019PGFA,he2020guided}.
	
	We can see that our proposed PGFL-KD achieves the competitive results on the holistic person ReID datasets. It is mentioned that our model uses only the vanilla ResNet model in testing, which does not introduce additional computational complexity and does not need a pose estimator. % (ISP \cite{zhu2020identity} needs human semantic parsing). 

	\subsection{Ablation Studies}

	In this section, we conduct ablation studies to evaluate the effectiveness of designs in the proposed PGFL-KD. PGFL-KD consists of a main branch (MB), and two pose-guided branches, \ieno, a foreground-enhanced branch (FEB), and a body part semantics aligned branch (SAB). Occluded-Duke is a larger occluded dataset, which can better reflect the effectiveness of the models. Table \ref{tab:module-analysis} shows the results. Model-1 denotes our Baseline, where ResNet50 network is trained followed by ReID loss.
	We denote whether SAB/FEB is enabled (denoted by On) or not (denoted by Off) in training in the column titled by $\mathcal{S}$ (means Switch). For all these schemes, the global feature $\boldsymbol f_G$ of the MB is used for testing.

	\noindent\textbf{Effectiveness of SAB.} %TODO
	As shown in Table \ref{tab:module-analysis}, we denote our interaction based optimization in SAB as $\mathcal{I}$, and multi-part contrastive loss (for knowledge distillation) in SAB as $\mathcal{M}$ (see Section \ref{subsec:SAB}). 
	%Firstly, when removing all proposed modules, our framework degrades to a baseline model~\cite{zheng2016person} (Model-1 in the first row), where only a main branch with global feature output $\boldsymbol f_G$ is available. Its performance is $52.7\%$ in Rank-1 accuracy.
	We denote Model-2 (MB+SAB) as a scheme when we add the SAB without $\mathcal{I}$ and $\mathcal{M}$, where the part semantics aligned feature $\boldsymbol f_P$ is followed by ReID loss. Then the SAB plays a role of regularizing the backbone feature learning. 
	%We use the global feature $\boldsymbol f_G$ for testing. 
	We can see that Model-2 outperforms Baseline by 0.6\%/1.3\% in mAP/Rank-1.
	
	When the interaction based optimization $\mathcal{I}$ of SAB is used, \ieno, Model-3 (MB+SAB w/ $I$), the performance is further improved by $\textbf{1.6\%/2.4\%}$ in mAP/Rank-1 in comparison with Model-2 (MB+SAB). This demonstrates the effectiveness of our proposed interaction-based training in promoting the semantics alignment for the global feature. 
	%strategy $\mathcal{I}$ for the semantic feature alignment is important for the occluded ReID. 
	In Model-4 (MB+SAB w/ $I M$), the using of the proposed multi-part contrastive loss ($\mathcal{M}$) explicitly enhances the channel-wise feature alignment of the global feature guided by the local part features of the SAB, which brings  additional \textbf{$3.9\%/3.0\%$} gain in mAP/Rank-1.

	\begin{table}[t]
	    \small
		\begin{center}
			\scalebox{1}{
				\centering
				\begin{tabular}{c|ccc|cc|cc}
					\hline
					\hline
					\multicolumn{1}{c|}{\multirow{2}{*}{Index (Scheme)}} & \multicolumn{3}{c|}{SAB}   & \multicolumn{2}{c|}{FEB}  &  \multicolumn{1}{c}{\multirow{2}{*}{R1}} & \multicolumn{1}{c}{\multirow{2}{*}{mAP}}     \\ 
			        \multicolumn{1}{c|}{}         & \multicolumn{1}{c}{$\mathcal{S}$}   & \multicolumn{1}{c}{$\mathcal{I}$} & \multicolumn{1}{c|}{$\mathcal{M}$}     & \multicolumn{1}{c}{$\mathcal{S}$} & \multicolumn{1}{c|}{ $\mathcal{C}$}  & \multicolumn{1}{c}{}   & \multicolumn{1}{c}{}   \\ \hline
				% 	Index & $\mathcal{I}$ & $\mathcal{M}$& $\mathcal{F}$ & $\mathcal{C}$ & Rank-1 & mAP \\ \hline
					1 (Baseline) & Off     & $\times$  & $\times$      & Off &$\times$ & 52.7 & 45.9 \\
					\hline
					2 (MB+SAB) & On & $\times$     & $\times$    &Off &$\times$ & 54.0 & 46.5 \\ 
					3 (MB+SAB w/ $I$) & On & $\checkmark$     & $\times$    &Off &$\times$ & 56.4 & 48.1 \\ 
					4 (MB+SAB w/ $I M$) & On & $\checkmark$   & $\checkmark$ &Off &$\times$ & 59.4 & 52.0 \\ 
					\hline
					5 (MB-SAB+FEB) & On & $\checkmark$ & $\checkmark$ & On &$\times$ & 61.2 & {52.1} \\
					6 (MB-SAB+FEB w/~$C$) & On & $\checkmark$ &  $\checkmark$ & On &$\checkmark$ & \textbf{63.0} & \textbf{54.1} \\
					\hline
					\hline
				\end{tabular}
			}
		\end{center}
		\caption{Effectiveness of our designs in the proposed PGFL-KD on Occluded-Duke. It consists of a main branch (MB), and two pose-guided branches, \ieno, a foreground-enhanced branch (FEB), and a body part semantics aligned branch (SAB).
		We denote the interaction-based otimization in SAB by $\mathcal{I}$, multi-part contrastive loss in SAB by $\mathcal{M}$, and consistent loss in FEB by $\mathcal{C}$. Note that for all these schemes, the global feature $\boldsymbol f_G$ of the MB is used for testing.}
		\vspace{-6mm}
		\label{tab:module-analysis}
	\end{table}
	
	\noindent\textbf{Effectiveness of FEB.} %TODO
	%As shown in Table \ref{tab:module-analysis}, we denote the consistent loss in FEB as $\mathcal{C}$. 
	We denote the consistent loss (for knowledge distillation) in FEB as $\mathcal{C}$.
	On top of Model-4 (MB+SAB w/ $I M$), when adding the FEB without $\mathcal{C}$, we denote the scheme as Model-5 (MB-SAB), where the foreground-enhanced feature $\boldsymbol f_E$ is followed by ReID loss. In this case, the FEB regularizes the feature learning of the backbone network. The performance is significantly improved by 0.1\%/1.8\% in mAP/Rank-1 over Model-4 (MB+SAB w/ $I M$). The foreground-enhanced operation in FEB intends to emphasise the features of visible body parts while alleviating the interference of obstructions and background. When we explicitly distilling the knowledge from the FEB to the MB by adding consistent loss (\ieno, $\mathcal{C}$ enabled), we can see that Model-6 (MB-SAB + FEB w/ $C$) is much superior than Model-5 that without using  consistent loss.
	Model-6 represents our final scheme PGFL-KD. Thanks to our designs, it outperforms Baseline significantly by \textbf{8.2\%/10.3\%} in mAP/Rank-1. 
	
	%By adopting our designed foreground-enhanced operation $\mathcal{F}$ in FEB, the performance is significantly improved by $\textbf{2.0\%}$ in Rank-1. The foreground-enhanced operation in FEB intends to emphasise the features of visible body parts while excluding the interference of obstructions and background.
	%To get rid of the dependency on pose information when testing, we regularize the MB to learn the merits of the FEB through knowledge distillation, where our PGFL-KD achieves the best accuracy of $63.0\%$ in Rank-1.

	\noindent\textbf{Effectiveness of Multi-part Constrastive Loss vs. Consistent Loss for the SAB.} To distill knowledge from the body part semantics aligned feature $\boldsymbol f_P$ in the SAB to the global feature $\boldsymbol f_G$ in the MB, we use multi-part constrastive loss ($\mathcal{L}_{MCL}$ ) for better alignment. Table~\ref{tab:partialreid_result_pose} shows that replacing this contrastive loss with a consistent loss $\mathcal{L}_{CL}$ (similar to Eq.(\ref{eq:loss_MMD}) in SAB, there is a 3.9\% drop in mAP accuracy.

	\begin{table}[t]%[h]
		\centering
		\begin{tabular}{c|cc}
			\hline 
			
			\hline 
			\multicolumn{1}{c|}{Method} & \multicolumn{1}{c}{Rank-1} & \multicolumn{1}{c}{mAP} 
			\\ \hline 
			
% 			{PGFL-KD w/o $\mathcal{L}_{ReID}^E$ }      & 59.4 & 51.7 \\
% 			{PGFL-KD w/o $\mathcal{L}_{CL}$ }              &  {61.2} & {52.1}\\
% 			{PGFL-KD w/o $\mathcal{L}_{ReID}^V$ }  &  {58.2} & {49.7}\\
% 			{PGFL-KD w/o $\mathcal{L}_{MCL}$ }              &  {60.4} & {50.9}\\
			MB + FEB + SAB w $\mathcal{L}_{CL}$   &  {59.1} & {50.0}\\
			MB + FEB + SAB w $\mathcal{L}_{MCL}$  &  63.0 &54.1\\
			\hline   
			\hline
		\end{tabular}
		%\vspace{3mm}
		\caption{Effectiveness of using different knowledge distillation losses for the SAB in our PGFL-KD on Occluded-Duke.}
		%Evaluation of the
		%PGFL-KD (Ours) achieves best performance.
		\vspace{-6mm}
		\label{tab:partialreid_result_pose}
	\end{table}

% 	\noindent\textbf{Analysis of the Loss Functions.}
% 	We conduct ablation studies to evaluate the effectiveness of loss functions in our PGFL-KD on Occluded-duke dataset, as shown Table~\ref{tab:partialreid_result_pose}. 
% 	By replacing the contrastive loss with a consistent loss in SAB, there is a 3.9\% drop in performance in mAP accuracy. 
% 	Moreover, $\mathcal{L}_{MCL}$ and $\mathcal{L}_{ReID}^V$ bring 3.2\% and 4.4\% performance improvements in term of mAP, respectively. It indicates that our multi-part contrastive loss $\mathcal{L}_{MCL}$ and interaction-based training $\mathcal{L}_{ReID}^V$ facilitate our model's ability to learn semantics aligned representations.
% 	$\mathcal{L}_{ReID}^E$ and consistent loss $\mathcal{L}_{CL}$ brings 2.4\%/2.0\% gain in Rank-1, which indicates these losses can explicitly alleviate the interference from the obstructions by learning foreground-enhanced feature in the FEB.

	\begin{figure}[t]%[h]
	\centering
	\small
	\includegraphics[width=0.48\textwidth]{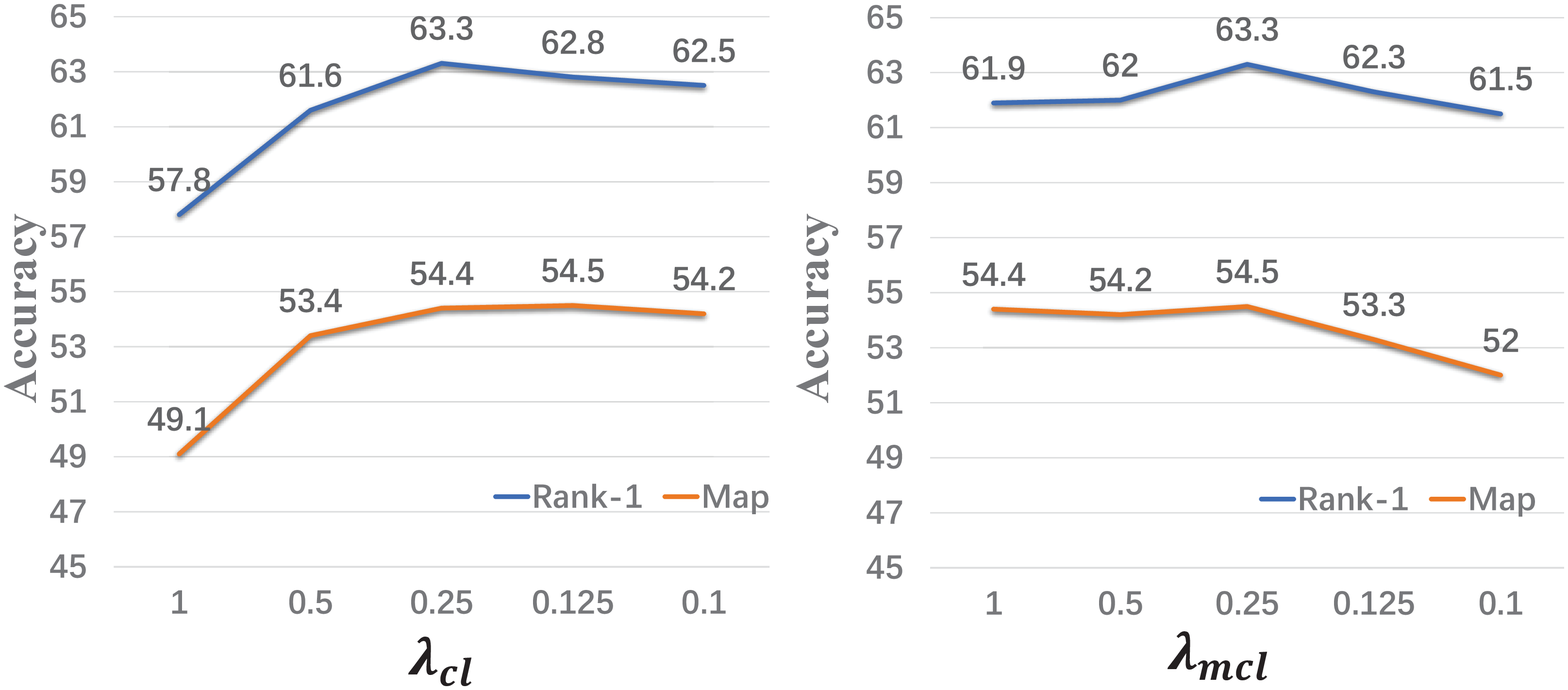}
	\caption{Evaluation of the proposed PGFL-KD with different values of parameter on Occluded-Duke. (a) multi-part contrastive loss $\lambda_{mcl}$ ;(b) consistent loss $\lambda_{cl}$.}
	\label{fig:hp}
	\end{figure}

	\begin{table}[t]%[h]
	\begin{center}
   % \samll
    \footnotesize
		\begin{tabular}{c|cc|cc}
			\hline
			\hline
			Feature for Testing& FLOPS & Param & Rank-1 & mAP \\ 
			\hline
			Global Feature $\boldsymbol{f}_G$ (Baseline)& 8.98G & 39.89M &52.7 & 45.9 \\
			Global Feature $\boldsymbol{f}_G$ (PGFL-KD) & 8.98G & 39.89M& 63.0 & 54.1 \\
			\hline
			Body Part Feature $\boldsymbol{f}_P$ &24.75G &116.09M & 57.1 & 46.7 \\
			Body Part Fused Feature $\boldsymbol{f}_V$  &24.75G &117.54M & 62.8 & 53.0 \\
			\hline
			Foreground Feature $\boldsymbol{f}_F$  & 24.72G&103.51M & 58.3 & 47.3 \\
			Foreground Fused Feature $\boldsymbol{f}_E$& 24.72G& 104.95M & 63.4 & 54.6 \\
			\hline
			\hline
		\end{tabular}
	\end{center}
	\caption{Performance (\%) and \tcbb{inference complexity} comparisons when we use different features for matching for our PGFL-KD on \tcbb{Occluded-Duke}.}
	\vspace{-3mm}
	\label{tab:feature-analysis}
		
	\end{table}

	\noindent\textbf{Influence of Different Hyper-parameters.}
	We study the influence of different hyper-parameters on the performance. Figure~\ref{fig:hp} shows the results. We can see that when $\lambda_{cl}=0.25$ and $\lambda_{mcl}=0.25$, PGFL-KD presents the best performance (in mAP).
	
	\subsection{Different Features for Matching and Inference Complexity}
	%\noindent\textbf{Different Features for Matching and Complexity.} 
	For our PGFL-KD scheme, we compare the performance when we use different features for matching in inference, and show the results in Table \ref{tab:feature-analysis}. 
	%For part-guided channel-wise learning network (PGFL-KD), we have three branches which focus on global features, body part features and foreground feature, respectively. 
	1) When we use the global feature $\boldsymbol{f}_G$ in testing, ours significantly outperforms Baseline by 8.3\%/10.3\% in term of mAP/Rank-1. This is only 0.5\%/0.4\% inferior to the best performance in term of mAP/Rank-1 which need to use pose information in testing. Through pose-guided interaction learning (\ieno, knowledge distillation and interaction-based training), we get rid of the dependency on the pose estimator, retaining high performance and low computational complexity in the test. The computational complexity is the same as Baseline, which is about 1/3 of the schemes which need a pose estimator.
	2) 	Body part feature $\boldsymbol{f}_P$ only or foreground feature $\boldsymbol{f}_P$ only is less effective since it lacks the global information. In contrast, $\boldsymbol{f}_G$ still preserves global information while inheriting the merits of pose-guided features. 
	3) The ensemble of the features ($\boldsymbol{f}_E$ and $\boldsymbol{f}_V$) further brings slight gain (about 0.5 in mAP). However, their computational complexity is about three times greater than that of using only the global feature.

	\begin{table}[t]%[h]
	\begin{center}
    % \footnotesize
		\begin{tabular}{cc|cc}
			\hline
			\hline
			Method & Time & Method & Time\\
			\hline
			DSR~\cite{he2018deep} & 4.84s & SFR~\cite{he2018recognizing} & 4.65s\\
			PGFA~\cite{miao2019PGFA}  & 0.82s & PGFA w/o pose~\cite{miao2019PGFA}   & 0.12s\\
            % PGFA+ w pose~\cite{miao2021identifying} & 0.85s & PGFA+ wo pose~\cite{miao2021identifying} & 0.14s\\
			PCB~\cite{sun2018beyond} & 0.09s & PGFL-KD (Ours) & 0.08s\\
			\hline
			\hline
		\end{tabular}
	\end{center}
	\caption{Inference speed (seconds per query) on Occluded-Duke.}
	\vspace{-3mm}
	\label{tab:feature-time}
	\end{table}
	
	%\tcr{
	We compare the inference speed of our method with PCB~\cite{sun2018beyond}, the partial re-id methods (DSR~\cite{he2018deep}, and SFR~\cite{he2018recognizing}), and the occluded re-id methods PGFA~\cite{miao2019PGFA}.
	%, as shown in Table~\ref{tab:feature-time}. 
	Table~\ref{tab:feature-time} shows that our method is much faster than other methods \tcbb{DSR~\cite{he2018deep}, and SFR~\cite{he2018recognizing} because there is no time-consuming feature map matching during inference in our method.
	%because our method only perform the vanilla baseline when testing. 
	Ours has similar inference speed with PGFA w/o pose \cite{miao2019PGFA} but achieves much better performance (see Table~\ref{tab:occluded_result})}
	%has similar speed with our
	%PGFA requires the extraction of pose landmarks, which is time consuming. Therefore, our method achieves faster inference by discarding pose branches. 
	%Our method is also much faster than the partial re-id methods (DSR and SFR) because there is no time-consuming feature map matching during inference in our method.
	%}

	\subsection{Feature Visualization} 
	
\begin{figure}[t]
	\centering
	\small
	\vspace{-3mm}
	\includegraphics[width=0.40\textwidth]{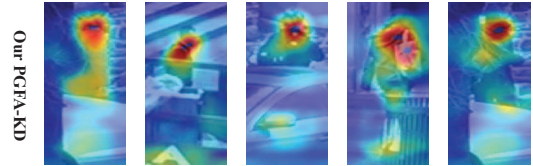}
	\caption{Visualization of the feature corresponds for our PGFL-KD scheme, where the responses for Baseline and the original images are shown in Figure~\ref{fig:intro}.}
	\vspace{-3mm}
	%The first row shows the samples on occluded duke, while the second and third rows show the feature corresponds for Baseline and our PGFL-KD with the guidance of pose .
	\label{fig:vis}
	\end{figure}

	As discussed in Section 3.2 and 3.3, we expect to let the main branch ignore the interference from obstructions/background and learn semantics aligned representations. We visualize the feature responses $F$ of our PGFL-KD in Figure~\ref{fig:vis}, where the responses for Baseline are shown in Figure~\ref{fig:intro}~(b). 
	In Figure~\ref{fig:intro}~(b), for the regions with objects occluding persons (\ieno, obstructions), the networks usually mistakenly generate high responses by regarding them as discriminative person regions.	With the guidance of FEB and SAB, the PGFL-KD focuses on the regions more related to foreground objects compared with Baseline. 
	%These regions play the most important role for person re-identification.
	
	\section{Conclusion}
	
	In this paper, we propose a network named Pose-Guided Feature Learning with Knowledge Distillation (PGFL-KD). PGFL-KD consists of a main branch (MB), a foreground-enhanced branch (FEB), and a body part semantics aligned branch (SAB). 
	Specifically, the FEB intends to emphasise the features of visible body parts while excluding the interference of obstructions and background (\ieno, foreground feature alignment).
	The SAB encourages different channel groups to focus on different body parts to have body part semantics aligned representation. 
	To get rid of the dependency on pose information and have a model of low complexity when testing, we regularize the main branch to learn the merits of the FEB and SAB through knowledge distillation and interaction-based training. 
	%In addition, the pose information is exploited to regularize the learning of semantics aligned features but is discarded in testing.
	Extensive experiments on occluded, partial, and holistic ReID tasks show the effectiveness of our proposed network and validate the superiority of PGFL-KD over various state-of-the-art methods.  
	   
	\section{Acknowledgments}
	
	This work was in part supported by the National Key R$\&$D Program of China under Grand 2020AAA0105702, National Natural Science Foundation of China (NSFC) under Grants U19B2038 and 61620106009, and China Postdoctoral Science Foundation Funded Project under Grant 2020M671898.

	{\balance
	\bibliographystyle{ACM-Reference-Format}
	\bibliography{egbib}}
	
\end{document}